\pdfoutput=1

\documentclass[11pt]{article}

\usepackage[preprint]{acl}

\usepackage{times}
\usepackage{latexsym}

\usepackage[T1]{fontenc}

\usepackage[utf8]{inputenc}

\usepackage{microtype}

\usepackage{inconsolata}

\usepackage{graphicx}
\usepackage{algorithm}
\usepackage{algpseudocode}
\usepackage{array}
\usepackage{multirow}
\usepackage{makecell}
\usepackage{booktabs}
\usepackage{bbding}
\usepackage{colortbl}
\usepackage{authblk}

\usepackage{footnote}

\title{Multimodal Cognitive Reframing Therapy via Multi-hop Psychotherapeutic Reasoning}

\author[1]{Subin Kim\thanks{Equal Contribution}}
\author[1]{Hoonrae Kim$^*$}
\author[1]{Heejin Do}
\author[1,2]{Gary Geunbae Lee}

\affil[1]{Graduate School of Artificial Intelligence, POSTECH, South Korea}
\affil[2]{Department of Computer Science and Engineering, POSTECH, South Korea}

\affil[ ]{\texttt{\{kimsubin, hoonrae, heejindo, gblee\}@postech.ac.kr}}

\begin{document}
\maketitle
\begin{abstract}
Previous research has revealed the potential of large language models (LLMs) to support cognitive reframing therapy; however, their focus was primarily on text-based methods, often overlooking the importance of non-verbal evidence crucial in real-life therapy. 
To alleviate this gap, we extend the textual cognitive reframing to multimodality, incorporating visual clues. 
Specifically, we present a new dataset called \textbf{M}ulti \textbf{M}odal-\textbf{Co}gnitive \textbf{S}upport \textbf{C}onversation (M2CoSC), which pairs each GPT-4-generated dialogue with an image that reflects the virtual client's facial expressions.
To better mirror real psychotherapy, where facial expressions lead to interpreting implicit emotional evidence, we propose a multi-hop psychotherapeutic reasoning approach that explicitly identifies and incorporates subtle evidence. Our comprehensive experiments with both LLMs and vision-language models (VLMs) demonstrate that the VLMs' performance as psychotherapists is significantly improved with the M2CoSC dataset. Furthermore, the multi-hop psychotherapeutic reasoning method enables VLMs to provide more thoughtful and empathetic suggestions, outperforming standard prompting methods.
\end{abstract}

\renewcommand{\arraystretch}{1.15}

\section{Introduction}
As a crucial part of cognitive behavioral therapy (CBT), \textit{cognitive reframing} addresses lots of mental health issues rooted in deeply ingrained negative and distorted thought patterns \cite{BECK1970184, Powles1974BeckAT, Beck1987CognitiveTA, Beck1988LoveIN,walen1992practitioner, halamandaris1997individual, ditomasso2000medical, hofmann2012efficacy}. 
Recently, studies attempted to utilize large language models (LLMs) in this task, highlighting their growing potential in the field of psychotherapy \cite{ziems-etal-2022-inducing, maddela-etal-2023-training, sharma-etal-2023-cognitive, qu2023conditioning, yang2023towards, 10.1145/3589334.3648137,  xiao-etal-2024-healme}.
Conventionally, cognitive reframing has been explored with text-based sentence rewriting methods aimed at shifting negative viewpoints to positive ones \cite{ziems-etal-2022-inducing, maddela-etal-2023-training, sharma-etal-2023-cognitive}.
Concerned that sentence-based cognitive reframing can feel unnaturally imposed, \citet{xiao-etal-2024-healme} suggest a three-stage conversational approach with LLMs encouraging clients to engage more actively and form self-positive viewpoints.

\begin{figure}[t!]
\centering
  \includegraphics[width=0.48\textwidth]{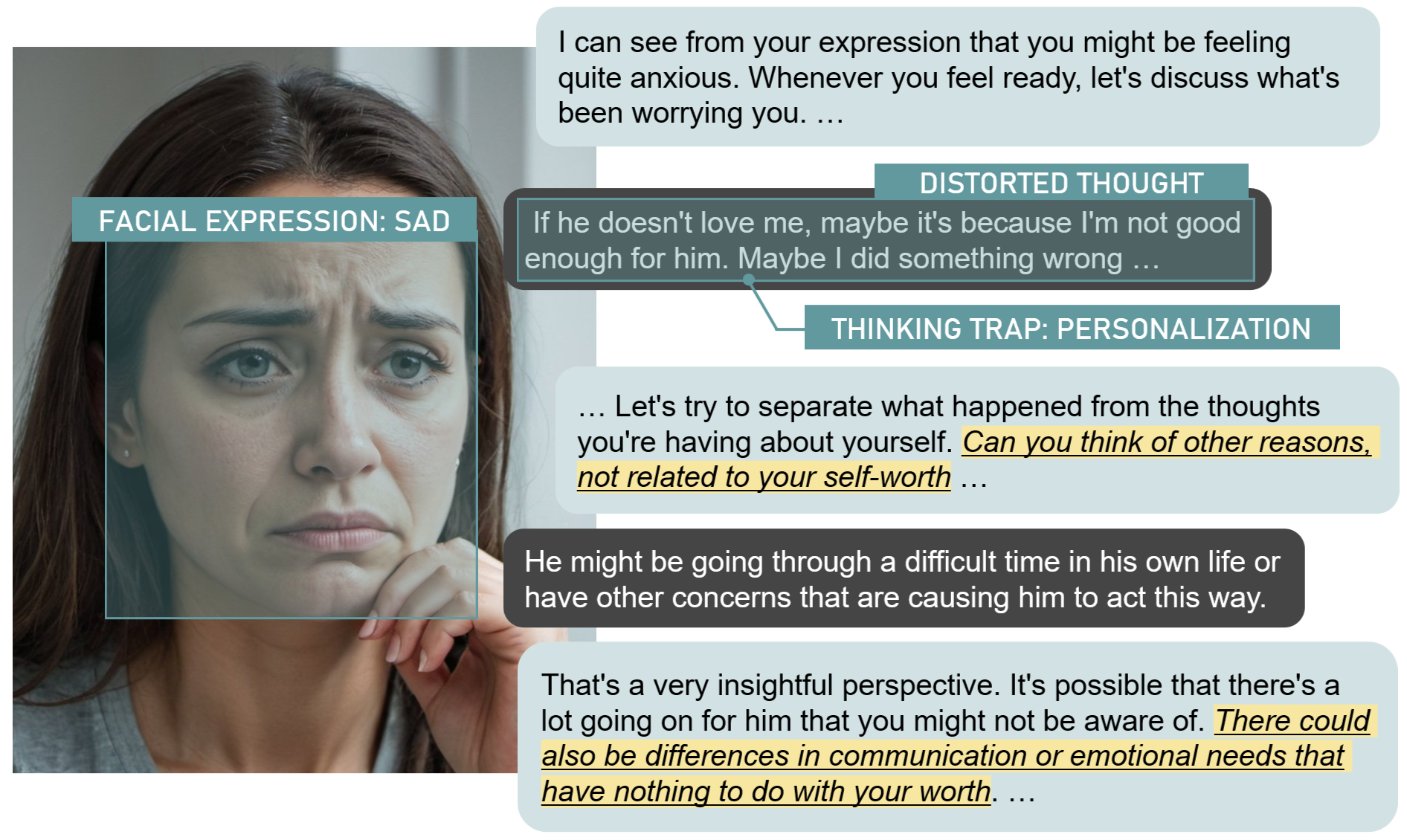}
  \caption{Illustration of a multimodal conversational cognitive reframing. The therapist uses both verbal and non-verbal information to assess the client's states and then provides appropriate interventions.\protect\footnotemark}
  \label{fig:mm_reframing}
\end{figure}

\footnotetext{To comply with the terms of the AffectNet license, all images presented in this paper are synthesized using DALL-E 3 \cite{betker2023improving} and not sourced from the AffectNet dataset.}

Despite the promising results of LLMs in previous systems, non-verbal aspects of psychotherapeutic theory are often overlooked, creating a significant gap between real face-to-face therapy and the systems. 
In actual psychotherapy situations, recognizing non-verbal emotions is essential for effective communication and is a critical skill closely linked to the therapist's ability to provide effective therapy \cite{10.1037/a0028807, Dollinger2021-cu}.

\begin{figure*}[t!]
  \includegraphics[width=\linewidth]{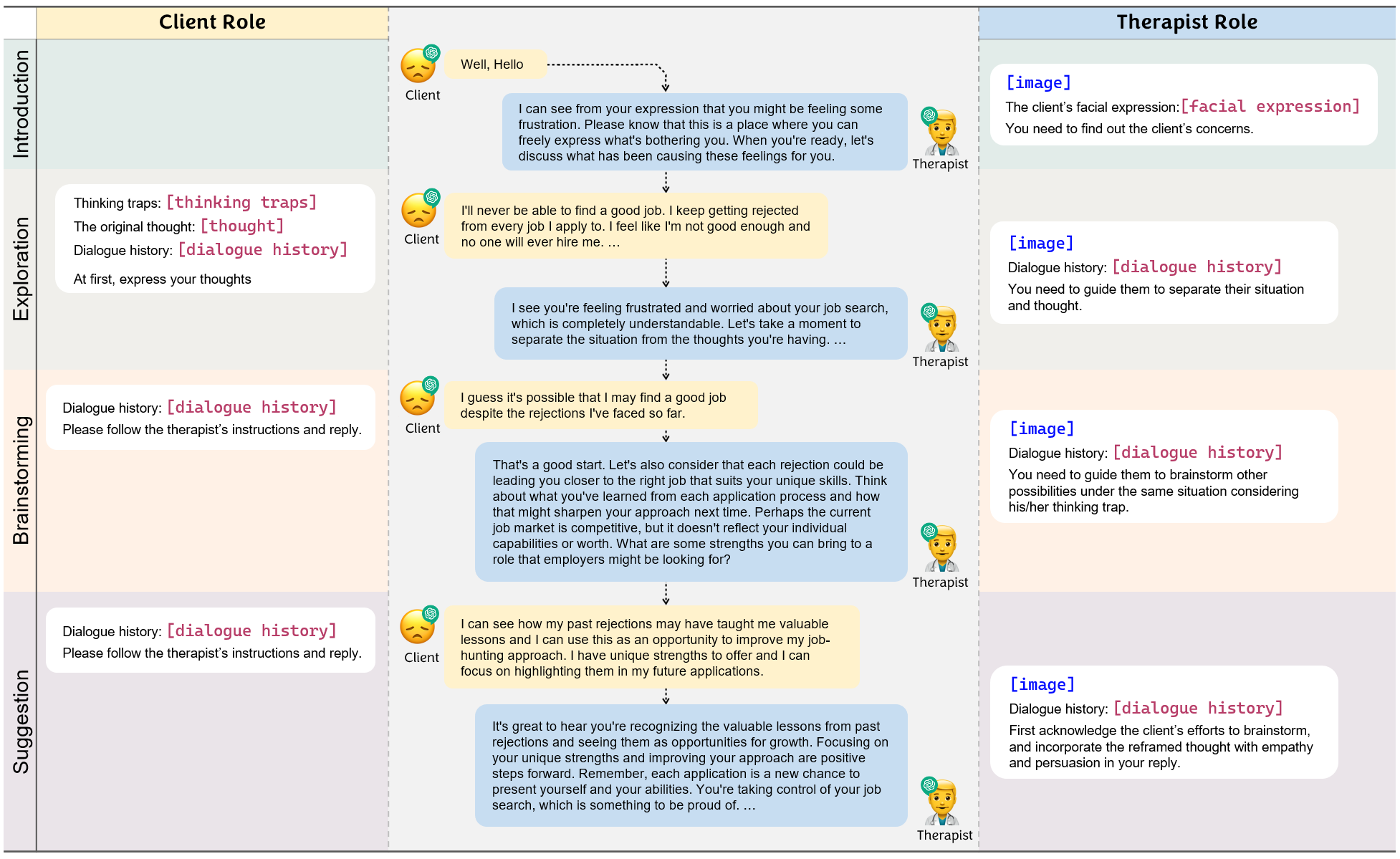}
  \caption{An example illustrating the construction of the M2CoSC dataset. Left: the prompt given to GPT-4 in the \textbf{client} role; Right: the prompt given to GPT-4 Vision in the \textbf{therapist} role. GPT-4 Vision is given a client's face image. \texttt{[dialogue history]} denotes a history of conversations accumulated during role play.
}
  \label{fig:dataset}
\end{figure*}

In this study, we propose to extend the concept of cognitive reframing into multimodality, integrating visual clues into the therapy process.
Our aim is to provide a natural and effective cognitive reframing framework within a multimodal context, incorporating clients' non-verbal clues in the conversation (Figure~\ref{fig:mm_reframing}).
To this end, we create a novel synthetic benchmark dataset, \textbf{M}ulti \textbf{M}odal-\textbf{Co}gnitive \textbf{S}upport \textbf{C}onversation (M2CoSC), which pairs each synthetic dialogue with an image of the client’s facial expression. 

To create M2CoSC, we utilize the powerful role-playing capabilities of LLMs. 
We leverage GPT-4 Vision and GPT-4 \cite{achiam2023gpt} to take on the roles of psychotherapist and client, respectively, simulating therapy sessions as shown in Figure~\ref{fig:dataset}. 
Inspired by counseling theory's \cite{krishnan2015counselling, thepsylog2024, claibournecounseling2024} \textit{Initial Disclosure} stage among the five stages, we add the Introduction phase into our multimodal psychotherapy framework, extending the three-stage model proposed by \citet{xiao-etal-2024-healme}: Introduction, Problem Exploration, Brainstorming, and Suggestion.
In addition, to provide explicit guidance based on the client’s state, we introduce a multi-hop psychotherapeutic reasoning method. 
By exploring the implicit evidence necessary for the therapy and then generating responses based on this evidence, the AI therapist can offer adequate guidance after identifying the client's state.

We evaluate our approach by conducting extensive experiments with two test scenarios, dialogue-level evaluation and stage-level evaluation, using both LLMs and Vision-Language Models (VLMs).
The results show that training with our M2CoSC significantly enhances the counseling capabilities of VLMs, surpassing those of existing LLMs. 
Moreover, the multi-hop psychotherapeutic reasoning method allows VLMs to offer more rational and empathetic suggestions, outperforming standard prompting methods. 
Human evaluations further suggest that capturing the client's facial expressions in the system, as practical therapists do, can remarkably assist counseling.

To sum up, our contributions are as follows: (1) We propose multimodal cognitive reframing therapy using non-verbal information, creating the M2CoSC dataset that pairs dialogues with client facial expressions. (2) We establish a baseline for the M2CoSC dataset and propose a multi-hop psychotherapeutic reasoning approach to improve the capabilities of VLMs in delivering rational therapeutic interventions.

This work is a first step toward bringing multimodal cognitive reframing into  AI-enhanced psychotherapy. By introducing a benchmark dataset and a structured reasoning approach, we hope to inspire future research on leveraging non-verbal cues for more effective therapeutic conversations.

\section{Problem Definition and Goals}
In cognitive reframing therapy, a therapist must understand the client's states, which include their problematic situations, distorted thoughts, and thinking traps. The therapist then encourages the client to consider alternative possibilities. Building rapport with the client by expressing empathy is also crucial \cite{horvath1993role, lambert2001research}. 
In the real-world psychotherapy procedure, these stages involve both verbal and non-verbal information, where the therapist has sufficient ability to understand the client's states.

Here, our goal is to enhance the abilities of an AI psychotherapist by leveraging non-verbal information, guiding it to focus on facial expressions and to comprehend the client's states.
Given the client's facial image and dialogue history, we aim to provide empathetic responses while maintaining a consistent focus on the client's issues throughout the procedure, offering rational interventions free from logical errors or contradictions.

To achieve our goals, we created the M2CoSC dataset founded on three key values—empathy, logical coherence, and guidance—which serve as evaluation criteria in a prior study \cite{xiao-etal-2024-healme}.

\begin{itemize}
    \item Empathy reflects the therapist’s ability to understand and connect with the client’s emotions, assisting in building trust, connection, and emotional support, all critical to a strong therapeutic relationship.
    \item Logical coherence denotes the therapist’s ability to organize thoughts and provide well-structured insights, enhancing the quality of the conversation.
    \item Guidance indicates the therapist’s capacity to offer practical advice, solutions, and direction, aiding the client to navigate challenges, make informed decisions, and achieve positive outcomes.
\end{itemize}

We also utilize overall scores encompassing all three items. (see Appendix \ref{sec:overall_trait} for details.)

\section{M2 Cognitive Support Conversation}

\subsection{Dataset Construction}
Actual therapy conversations are rarely accessible due to the sensitive nature of mental health therapy; thus, we have created a synthetic dataset that can be shared with the research community\footnote{Under AffectNet's license, M2CoSC dataset is partially available at \url{https://github.com/nobel-postech/M2CoSC}.}. 
To construct multimodal conversational cognitive reframing dataset, we utilize two publicly available sources: the Facial Expression Recognition (FER) dataset called AffectNet \cite{8013713}, and the cognitive reframing dataset from \citet{sharma-etal-2023-cognitive}.
To address potential privacy concerns associated with using images of real people from AffectNet, we obtained agreement for all research participants, ensuring full compliance with AffectNet's policies.

For construction, we set up role-play scenarios with two agents: GPT-4 in the client role and GPT-4 Vision in the therapist role.

As shown in Figure \ref{fig:dataset}, we prompt GPT-4 in the client role and GPT-4 Vision in the therapist role using a set of four inputs: \textit{image, facial expression, thinking traps}, and \textit{thought}. \footnote{We used version {\fontfamily{qcr}\selectfont gpt-4-0613} of the GPT-4 API and version {\fontfamily{qcr}\selectfont gpt-4-1106-vision-preview} of the GPT-4 Vision API.}
The \textit{image} represents the client's facial image, the \textit{facial expression} denotes the client's facial expression, the \textit{thought} reflects the client's thoughts, and the \textit{thinking traps} capture cognitive distortions present in the \textit{thought}.

For facial expressions and images, We employ AffectNet, containing publicly accessible images from the internet collected for research under non-commercial use. 
For thinking traps and thoughts, we utilize the well-designed open-sourced dataset from \citet{sharma-etal-2023-cognitive}, which was collected following ethical guidelines, including informed consent and participant privacy safeguards.
To the best of our knowledge, this work is the first to combine multiple datasets to create a multimodal conversation specifically designed for the mental health domain.

Each dialogue consists of four turns, which correspond to different stages of a psychotherapeutic conversation. 
In this context, a "turn" is the same as a "stage."
The action expected from the client is to follow the psychotherapist's instructions, and the actions required for the psychotherapist at each stage are as follows.
\begin{enumerate}
    \item \textbf{Introduction}: The AI psychotherapist expresses empathy and encourages the client to explore their problems.
    \item \textbf{Exploration}: The AI psychotherapist guides the client to distinguish their thoughts from their situation.
    \item \textbf{Brainstorming}: The AI psychotherapist discusses other possibilities for the client's interpretation. This involves asking about the basis for the client's thoughts or considering the possibility of alternative interpretations.
    \item \textbf{Suggestion}: The AI psychotherapist first recognizes the client's effort to explore other possibilities and presents specific and rational suggestions for the client.
\end{enumerate}

Considering the characteristics of cognitive reframing counseling, which often involves addressing negative emotions, we excluded the "happy" expression from the 8 facial expressions in AffectNet. The matching between (\textit{image, facial expression}) and (\textit{thinking traps, client's thought}) was performed randomly with uniform distribution.

The statistics for the M2CoSC dataset are summarized in Table \ref{tab:statistics}. 
The M2CoSC dataset contains a total of 429 conversations, each consisting of exactly four turns.

\begin{table}[t]
  \centering
  \resizebox{\columnwidth}{!}{%
  \begin{tabular}{cc|cc|c}
    \toprule
    &  & \multicolumn{2}{c|}{\textbf{Avg. Tokens}}  \\
    & \textbf{\# of Dialogue}& \textbf{Client} & \textbf{Therapist} & \textbf{Rounds} \\
    \midrule
    Train    & 329  &  24.93  & 63.64 & 4    \\
    Test     & 100  &  24.01  & 62.81 & 4    \\
    \bottomrule
  \end{tabular}
  }
  \caption{\label{tab:statistics}
    Dataset statistics for M2CoSC. \textit{\# of Dialogue} indicates the total number of dialogues in the subset. \textit{Avg. Tokens} represents the average number of tokens per utterance from the \textit{Client} and the \textit{Therapist}. \textit{Rounds} denotes the number of turns per dialogue in the subset.
  }
\end{table}

\begin{table}[b]
  \centering
  \resizebox{0.75\columnwidth}{!}{%
  \begin{tabular}{cc}
    \toprule
    & \textbf{Image-Dialogue Consistency} \\
    \addlinespace[-3pt]
    & {\small (0-2)} \vspace{0.1em}\\
    \hline
    Train    & 1.472    \\
    Test     & 1.667    \\
    \bottomrule
  \end{tabular}
  }
  \caption{\label{tab:image_dial_consistency}
    \textit{Image-Dialogue Consistency} on the M2CoSC dataset.
  }
\end{table}

\subsection{Dataset Cleansing}
\label{sec:data_cleansing}
To ensure the quality of the M2CoSC dataset, we conducted manual data cleansing with the three native speakers, focusing on four aspects: \textit{Client-clarity}, \textit{Client-role}, \textit{Therapist-role}, and \textit{Image-Dialogue Consistency} (see Appendix \ref{sec:cleansing_manual} for detailed criteria).
The \textit{Image-Dialogue Consistency} is a criterion that evaluates whether the client's visual information and dialogue are consistent. 
If any of the four criteria received a score of 0, the corresponding data was deleted.
Table \ref{tab:image_dial_consistency} indicates a considerable correlation between the client's facial expressions and their verbal responses in our M2CoSC dataset.

We hired three native English speakers through Upwork\footnote{\url{https://www.upwork.com/}}, a crowdsourcing platform, to support this cleansing process.

\subsection{Dataset Quality Validation}
\label{sec:data_quality_assurance}

To further validate the quality of the M2CoSC, we evaluate the test set of the M2CoSC dataset based on three criteria: empathy, logical coherence, and guidance, along with an overall score. 
Each criterion was rated on a scale from 0 to 3 following the manual provided by \citet{xiao-etal-2024-healme}.
The test set evaluation was conducted using both human and GPT-4 assessments.
We hired an English-fluent psychotherapist for this evaluation and engaged them to manually evaluate the test set according to the detailed guidelines. 
We also employed GPT-4 for evaluation, feeding it the scorecard criteria and the dialogues from the test set.

The results in Table \ref{tab:testset_eval} indicate that both the human evaluator and GPT-4 provided high scores, showing similar scoring tendencies except for "Guidance." 
GPT-4 tended to give lower scores in this aspect, possibly attributable to its higher degree of expectation for problem-solving than human evaluators \cite{chiu2024computational}. 
Despite this discrepancy, the overall consistency in the other criteria supports the dataset's reliability.

\begin{table}[t]
  \centering
  \resizebox{\columnwidth}{!}{%
  \begin{tabular}{cc>{\centering\arraybackslash}m{2cm}cc}
    \toprule
    & \textbf{Empathy} & \textbf{Logical Coherence}  & \textbf{Guidance} & \textbf{Overall}\\
    \addlinespace[-3pt]
    & {\small (0-3)} & {\small (0-3)} & {\small (0-3)} & {\small (0-3)} \\ 
    \midrule
    Human  & 2.929  &  2.980  & 2.879  & 2.929  \\
    GPT-4  & 2.920  &  2.930  & 2.400  & 2.420 \\
    \bottomrule
  \end{tabular}
  }
  \caption{\label{tab:testset_eval}
    Human and GPT-4 evaluation results on M2CoSC test set. 
  }
\end{table}

\begin{figure*}[t!]
  \includegraphics[width=\linewidth]{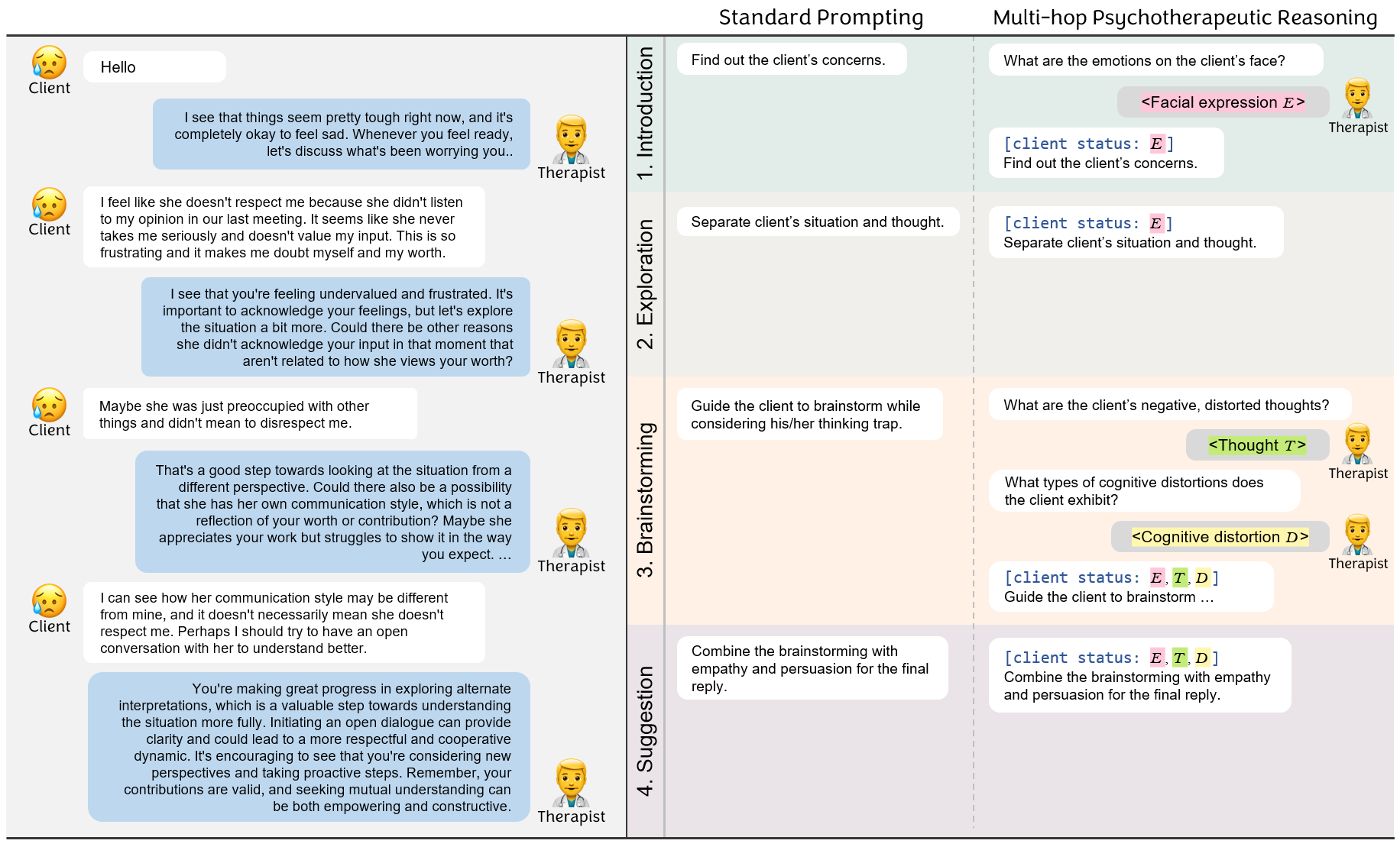}
  \caption{Comparison of standard prompting and multi-hop psychotherapeutic reasoning. 
  The multi-hop approach integrates the client's emotional and cognitive state (facial expressions, thoughts, and cognitive distortions) at each step of the intervention. The conversation on the left shows the therapist’s replies, which correspond to the four stages—Introduction, Guidance, Brainstorming, and Suggestion—outlined on the right.
  }
  \label{fig:reasoning}
\end{figure*}

\subsection{Multi-hop Psychotherapeutic Reasoning}
To ensure that the interventions are tailored to the client's needs, professional psychotherapists typically first understand the client's state and then provide interventions grounded on that \cite{greenberg1989emotion, rice1996facilitating}.
To mimic the real therapy process, we introduce multi-hop psychotherapeutic reasoning. 
This approach identifies implicit evidence crucial for cognitive reframing and incorporates it into step-by-step instructions. 
Initially, the AI therapist detects the client's state and then generates a response based on it, as illustrated in Figure \ref{fig:reasoning}.
In this work, we focus on three major aspects of the client's states—\textit{facial expression}, \textit{thought}, and \textit{thinking traps}—which accumulate over the stage of the conversation.
Each piece of evidence is identified at the appropriate stage. 
The detected evidence is included in the client's states, and the states are fed to the AI therapist as the prompt for the next evidence detection.

\section{Experiments}

\subsection{Settings}

\paragraph{Baseline models.} We utilize two representative models for our experiments: 
LLaMA2-chat-7b \cite{touvron2023llama}\footnote{\url{https://huggingface.co/meta-llama/Llama-2-7b-chat-hf}}, 
which is widely used in text generation tasks, and 
LLaVA-v1.5-7b\footnote{\url{https://huggingface.co/liuhaotian/llava-v1.5-7b}}, renowned for vision-related tasks.
For simplicity, we refer to LLaMA2-chat-7b as LLaMA2 and LLaVA-v1.5-7b as LLaVA in this work.
In addition, we denote the versions of LLaMA2 and LLaVA that were trained on the M2CoSC dataset as CS-LLaMA2 and CS-LLaVA, respectively. When multi-hop psychotherapeutic reasoning is applied, we add MH to their names.

\begin{table*}[t!]
  \centering
  \resizebox{0.8\textwidth}{!}{
  \begin{tabular}{c>{\raggedright\arraybackslash}m{3.2cm}c>{\centering\arraybackslash}m{2cm}ccc}
    \toprule
    & & \textbf{Empathy} & \textbf{Logical Coherence} & \textbf{Guidance} & \textbf{Overall} & \textbf{Avg.} \\ 
    \midrule\vspace{0.2em}
    \multirow{2}{*}{Baselines} & LLaMA2    & 2.665$^*$  &  2.390$^*$  & 1.600$^*$  & 1.540$^*$ & 2.218$^*$  \\
                            & LLaVA    & 2.640$^*$  &  2.570$^*$  & 1.790$^*$  & 1.740$^*$  & 2.333$^*$   \\
    \midrule\vspace{0.2em}
    \multirow{3}{*}{Ours} & CS-LLaMA2   & 2.690$^*$  &  2.410$^*$  & 1.640$^*$  & 1.580$^*$  & 2.247$^*$  \\
                        & CS-LLaVA      & 2.915$^*$  &  2.890     & 2.380     & 2.400     & 2.728     \\
                      \rowcolor[HTML]{dffcf0} \cellcolor{white} & CS-LLaVA w/ MH & \textbf{2.980} & \textbf{2.960} & \textbf{2.510} & \textbf{2.490} & \textbf{2.817} \\ 
 \bottomrule
  \end{tabular}
  }
  \caption{\label{tab:gpt_manual_aisim}
    Dialogue-level assessment results evaluated by GPT-4 using a role-playing approach with an AI client. Stars (*) next to values indicate a significant difference compared to \colorbox[HTML]{dffcf0}{CS-LLaVA w/ MH}, with a p-value < 0.05 determined by the paired t-test.   }
\end{table*}

\paragraph{Hyper-parameters.} Both LLaMA2 and LLaVA were fine-tuned with LoRA \cite{hu2022lora} on the M2CoSC dataset using default settings, except for the number of epochs. 
For LLaMA2, we used the official open-source models from Hugging Face, and for LLaVA, we followed the official code defaults\footnote{Default settings from \url{https://github.com/haotian-liu/LLaVA/tree/main}}. 
We split the M2CoSC train set into 80:20 training and validation subsets to select the optimal epoch based on validation performance. 
All models were trained with 4 $\times$ A100-80GB GPUs, using a batch size of 32 per GPU.

\setlength{\arrayrulewidth}{0.5pt}

\subsection{Evaluators}

\paragraph{GPT-4.} 
Recent research has shown that the evaluation of natural language generation (NLG) models using GPT-4 closely aligns with human evaluations. 
Therefore, GPT-4 is increasingly used as a judge for NLG tasks across various domains, including common applications, medical fields, and mathematics \cite{liu-etal-2023-g, sottana-etal-2023-evaluation, hsu-etal-2023-gpt, khondaker-etal-2023-gptaraeval, xiao-etal-2024-healme}.
Also, in conversation models, \citet{zheng2023judging} showed that GPT-4 achieves high agreement with human judgment in evaluations, releasing the corresponding judging prompt and the used codes\footnote{We utilize prompts from \url{https://github.com/lm-sys/FastChat/tree/main/fastchat/llm_judge}}. Building on this research, we evaluated the AI therapists using GPT-4 (API version)\footnote{We used the {\fontfamily{qcr}\selectfont
gpt-4-0613
} version of the GPT-4 API.} for evaluation in two ways:
\begin{itemize}
    \item \textbf{Score assessment}: We adopt a three-dimensional scoring system for AI therapists, evaluating them on empathy, logical coherence, and guidance.
    \item \textbf{Pairwise comparison}: We compared the interventions of therapists to determine whether Model A is better than Model B and vice versa or if it’s a tie for all possible pairs.\footnote{To ensure fairness and prevent position bias, we tested each case twice, swapping the positions each time.}
\end{itemize}\leavevmode
\paragraph{Human.} 
To enhance the reliability of the intervention evaluation, we conducted human evaluations with domain experts. 
We hired two fluent English-speaking psychotherapists via Upwork. 
The experts performed a pairwise comparison between our CS-LLaVA with multi-hop psychotherapeutic reasoning and others (see Appendix \ref{sec:inst_human_eval}).

\section{Results and Discussions}
For reliable comparison, we compared the performance of both LLMs and VLMs with two test scenarios: dialogue-level evaluation and stage-level evaluation.
The dialogue-level testbed, which has been used in prior research, allows us to observe how interventions are carried out throughout conversations. 
However, relying solely on this testbed makes it difficult to accurately compare the AI therapists' abilities due to the variability of the AI client.
To better assess interventions in terms of empathy, logical coherence, and rationality, we also conducted a stage-level evaluation.
This approach enabled us to compare therapists' interventions more precisely by analyzing turn-level performance on the M2CoSC test set, using consistent contextual input across the models.\\

\begin{figure}[t]
  \includegraphics[width=\columnwidth]{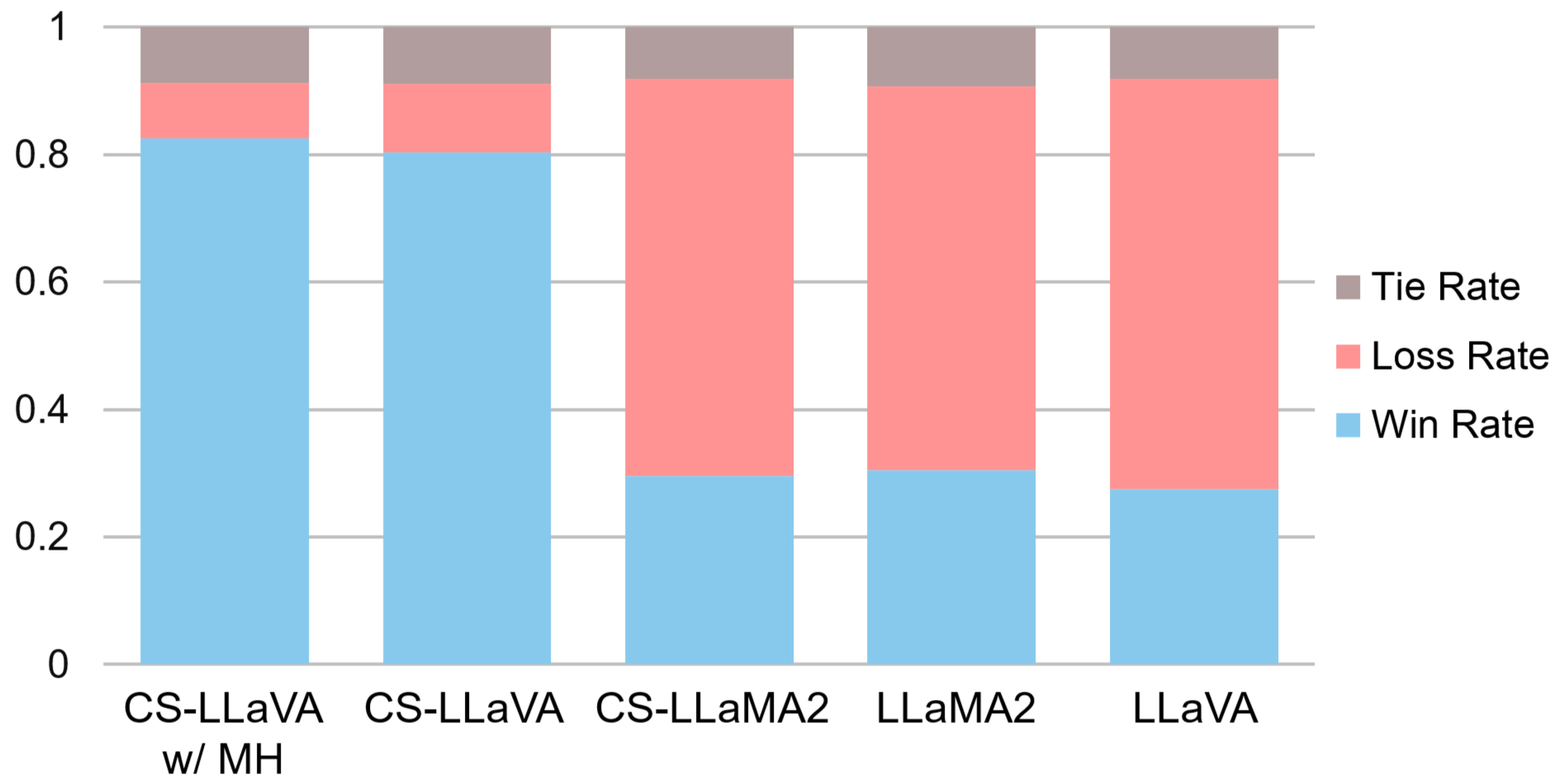}
  \caption{Dialogue-level win rates assessed by GPT-4. Detailed numerical results are provided in Appendix \ref{sec:gpt_wr_aisim}.}
  \label{fig:gpt_wr_aisim}
\end{figure}

\begin{table*}[t!]
  \centering
  \renewcommand{\arraystretch}{1.2}
  \resizebox{\linewidth}{!}{%
\begin{tabular}{c>{\raggedright\arraybackslash}m{3cm}|ccc|ccc|ccc|ccc}
    \toprule
    & & \multicolumn{3}{c|}{\textbf{Introduction}} & \multicolumn{3}{c|}{\textbf{Exploration}} & \multicolumn{3}{c|}{\textbf{Brainstorming}} & \multicolumn{3}{c}{\textbf{Suggestion}} \\ 
    & & \textbf{Emp.} & \textbf{Coh.} & \textbf{Gui.} & \textbf{Emp.} & \textbf{Coh.} & \textbf{Gui.} & \textbf{Emp.} & \textbf{Coh.} & \textbf{Gui.} & \textbf{Emp.} & \textbf{Coh.} & \textbf{Gui.} \\ 
    \midrule
    \multirow{2}{*}{Baselines} & LLaMA2 & 1.58$^*$ & 1.79$^*$ & 0.80$^*$ & 2.16 & 2.20$^*$ & 1.03$^*$ & 2.10$^*$ & 2.18$^*$ & 1.44$^*$ & 2.17$^*$ & 2.06$^*$ & 0.97$^*$ \\ 
        & LLaVA & 0.64$^*$ & 0.98$^*$ & 0.05$^*$ & 1.94$^*$ & 1.96$^*$ & 1.12$^*$ & 1.86$^*$ & 1.99$^*$ & 1.39$^*$ & 2.21$^*$ & 2.37$^*$ & 1.50$^*$ \\
    \midrule\vspace{0.1em}
    \multirow{2}{*}{Ours} & CS-LLaVA & 1.87$^*$ & 1.99 & 0.92$^*$ & 2.15 & 2.24$^*$ & \textbf{1.64} & 2.11$^*$ & 2.25$^*$ & 1.68 & 2.54 & 2.61 & 1.71 \\ 
      & \cellcolor[HTML]{dffcf0} CS-LLaVA w/ MH & \cellcolor[HTML]{dffcf0} \textbf{2.11} & \cellcolor[HTML]{dffcf0}\textbf{2.16} & \cellcolor[HTML]{dffcf0}\textbf{1.02} & \cellcolor[HTML]{dffcf0}\textbf{2.23} & \cellcolor[HTML]{dffcf0}\textbf{2.39} & \cellcolor[HTML]{dffcf0}1.60 & \cellcolor[HTML]{dffcf0}\textbf{2.27} & \cellcolor[HTML]{dffcf0}\textbf{2.39} & \cellcolor[HTML]{dffcf0}\textbf{1.79} & \cellcolor[HTML]{dffcf0}\textbf{2.59} & \cellcolor[HTML]{dffcf0}\textbf{2.67} & \cellcolor[HTML]{dffcf0}\textbf{1.80} \\ 
    \bottomrule
  \end{tabular}
  }
  \caption{\label{tab:gpt_manual_mmcsconv}
    Stage-wise assessment results as evaluated by GPT-4 at each stage on the M2CoSC testset. \textbf{Emp.}, \textbf{Coh.}, and \textbf{Gui.} represent Empathy, Logical Coherence, and Guidance, respectively. Stars (*) next to values indicate a significant difference compared to \colorbox[HTML]{dffcf0}{CS-LLaVA w/ MH}, with a p-value < 0.05  as determined by the paired t-test.
  }
\end{table*}

\begin{figure*}[t!]
  \includegraphics[width=0.97\textwidth]{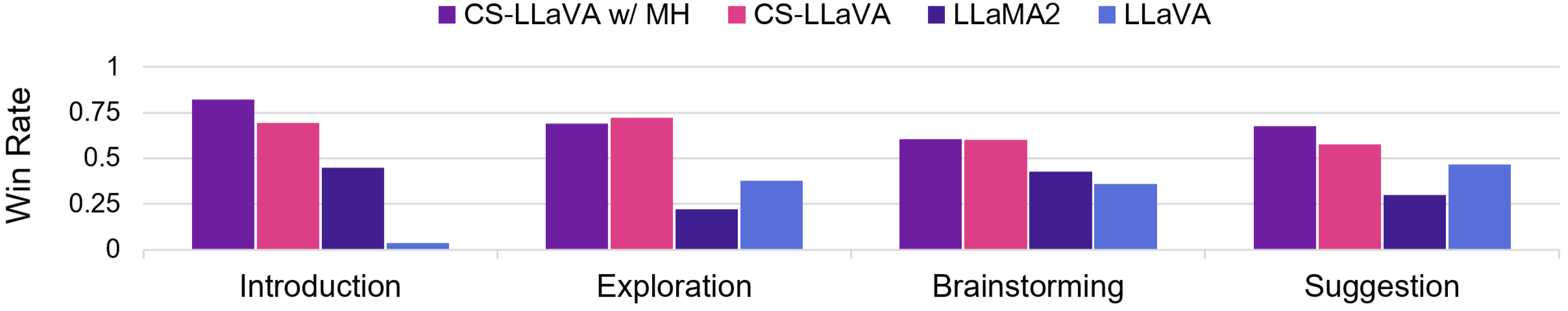}
  \caption{Stage-wise win rates assessed by GPT-4 at each stage of the M2CoSC benchmark. Numerical results are provided in Appendix \ref{sec:gpt_wr_csconv}.}
  \label{fig:gpt_wr_mmcsconv}
\end{figure*}

\subsection{Dialogue-level Evaluation}
In this scenario, we employ ChatGPT (API version)\footnote{We used the {\fontfamily{qcr}\selectfont
gpt-3.5-turbo-0125
} version of the ChatGPT API.} as an AI client to test our approach in AI-to-AI scenarios.
For prompting to AI client, we leverage 100 resources which are used as base resources from the test set, originally sourced from \citet{sharma-etal-2023-cognitive} and \citet{8013713}.
The AI client's role aligns with our data construction method, and we use the same prompts throughout the process.
To evaluate the performance difference between using only the text modality and incorporating image information, we also compared the results of CS-LLaMA2. 
For CS-LLaMA2, only the text modality was used without incorporating image information.

Table \ref{tab:gpt_manual_aisim} shows the dialogue-level assessment results evaluated by GPT-4.
Our M2CoSC dataset with the LLaVA family of models led to significant improvements across all aspects. 
By integrating multi-hop psychotherapeutic reasoning with three implicit evidences—\textit{facial expressions}, \textit{thoughts}, and \textit{thinking traps}—the models achieved enhancements across all evaluation aspects, with a particularly remarkable improvement in empathy.
These results support our hypothesis that understanding the client's emotional state before responding leads to more tailored and compassionate interactions.

LLaMA2, in contrast, shows minimal improvement when trained on the M2CoSC dataset, primarily due to the absence of visual information during the Introduction stage, which hampers effective training and results in subtle change.
These findings validate our hypothesis that models integrating multimodal information in counseling conversations possess superior therapeutic qualities compared to models relying solely on text. Consequently, we shifted our focus to the LLaVA family for further analysis instead of assessing CS-LLaMA2 in stage-level evaluations, except during human evaluation settings.

Similar trends were observed in the pairwise comparison (Figure \ref{fig:gpt_wr_aisim}). 
Despite showing the lowest performance, LLaVA exhibited a significant improvement when training on the M2CoSC and applying our multi-hop psychotherapeutic reasoning method, achieving the highest performance in CS-LLaVA w/ MH.
No significant performance difference was noted between LLaMA2 and CS-LLaMA2 due to the absence of visual information, further highlighting the impact of multimodal integration in counseling conversations.

\begin{figure}[t]
  \includegraphics[width=\columnwidth]{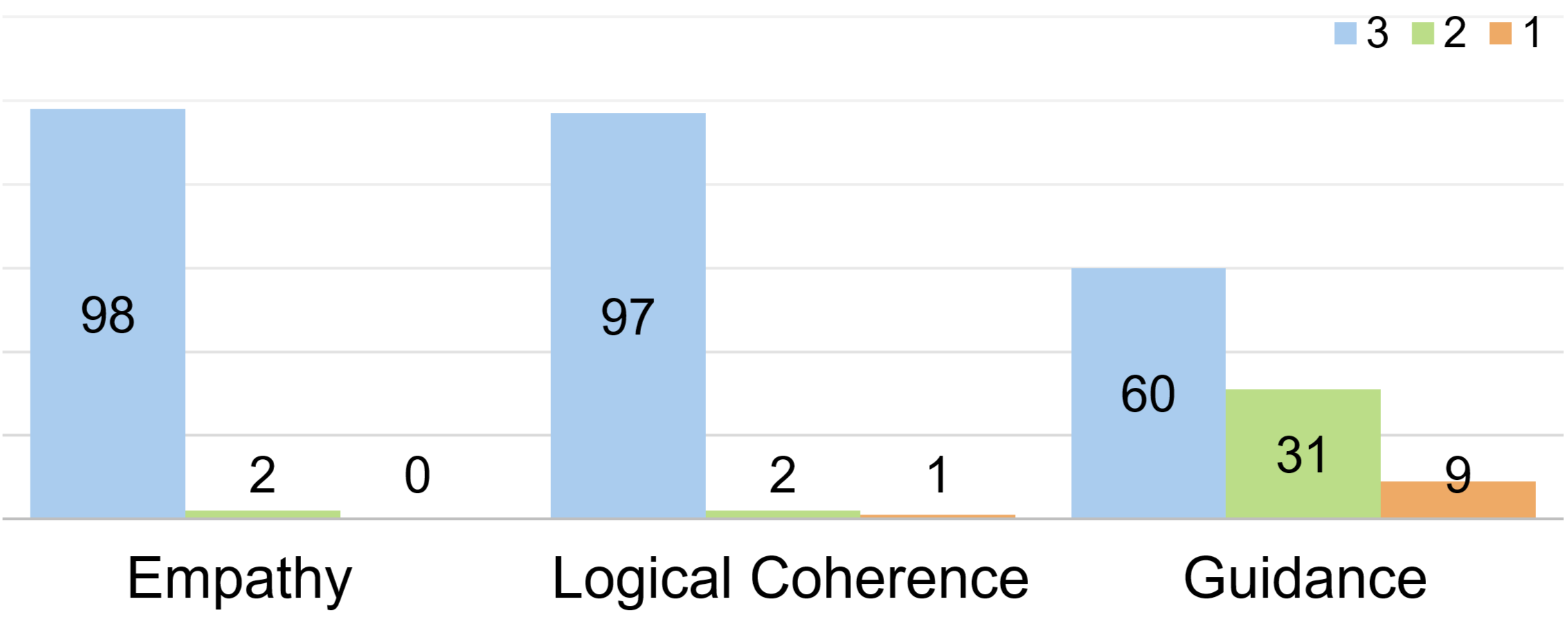}
  \caption{Dialogue-level score assessment ratio for CS-LLaVA w/ MH evaluated by GPT-4}
  \label{fig:dialog_level_ratio}
\end{figure}

\paragraph{Error analysis} 
Examining the aspect-wise results of our CS-LLaVA w/ MH, we found that the guidance received lower scores than the other aspects (Figure \ref{fig:dialog_level_ratio}). Hence, we further conducted a detailed error analysis, where the primary (six out of nine) cases were the AI therapist's failure to offer forward-looking strategies. Specifically, the AI therapist could not provide future-oriented strategies, hindering its ability to assist clients in preventing similar possible distortions. Detailed examples and additional qualitative analysis are described in Appendix \ref{sec:case_study} and \ref{sec:error_analysis}.

\begin{figure}[t!]
\centering
  \includegraphics[width=0.85\columnwidth]{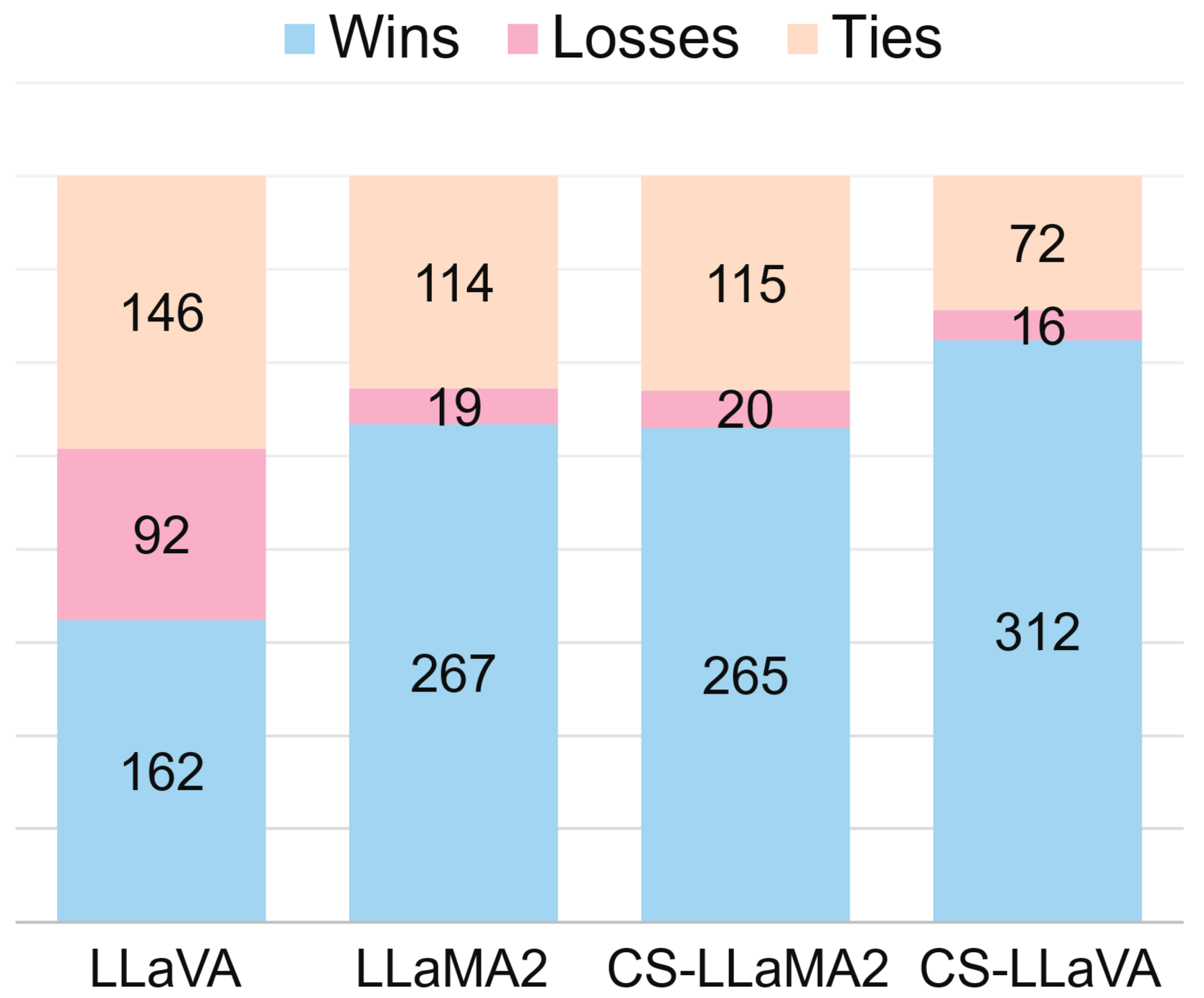}
  \caption{All stages aggregated win rates of CS-LLaVA w/ MH against other models, as evaluated by two psychotherapists on the M2CoSC test set. The domain experts are asked to choose which intervention is better at assessing the given intervention.}
  \label{fig:human_wr_mmcsconv}
\end{figure}

\subsection{Stage-level Evaluation}
In this testbed, each AI therapist responds to the same dialogue history, allowing us to directly compare their interventions.
To ensure reliability, we conducted evaluations using both GPT-4 and two human psychotherapists, and the evaluation was carried out at the turn level for each stage. As in the dialogue-level evaluation, we report score assessment and win rate results.

Throughout both assessment results and pairwise comparisons, the M2CoSC dataset had a noticeable impact across all stages.
Multi-hop psychotherapeutic reasoning outperformed the standard prompting method across most stages; however, in the Exploration stage, the guidance performance was slightly lower. 
This is likely because this stage emphasizes a deeper exploration of the client's situation rather than offering direct suggestions.

Stage-wise assessment results evaluated by GPT-4 (Table \ref{tab:gpt_manual_mmcsconv}) reveal that the score distribution is lower than that of the dialogue-level evaluation, as it assesses intervention at the turn level rather than the entire conversation. 
One key finding is the substantial performance improvement of the multi-hop psychotherapeutic reasoning process observed in the Introduction stage. The results demonstrate that our attempt to initially detect and incorporate the client's emotions led to more empathetic and coherent interactions. 
Figure \ref{fig:gpt_wr_mmcsconv} illustrates the stage-wise pairwise comparison results among the four models, as judged by GPT-4. Note that LLaVA, without our method, had the lowest win rate during the Introduction stage. This is attributable to its difficulty in expressing empathy using the client's non-verbal evidence, as it had not yet learned to effectively integrate multimodal information into conversations. 
The results highlight the importance of teaching models to recognize and utilize such information in counseling phases.

To further strengthen the reliability of the human evaluation results, we derived the win rate by comparing the proposed methodology with other approaches. 
Specifically, we compared CS-LLaVA with multi-hop psychotherapeutic reasoning to CS-LLaVA with standard prompting and other baselines, as evaluated by two domain experts. All stages aggregated results in Figure \ref{fig:human_wr_mmcsconv} exhibit that CS-LLaVA w/ MH achieved the highest wins and significantly fewer losses.
Stage-wise results in Table~\ref{tab:human_wr_mmcsconv} show that CS-LLaVA w/ MH outperforms across all stages. However, in the Exploration stage, CS-LLaVA performed similarly to CS-LLaVA w/ MH, likely due to the nature of the stage, which focuses more on exploring the client’s situation than providing suggestions.

Overall results strengthen the essence of integrating multi-hop psychotherapeutic reasoning, particularly in stages where understanding and responding to emotional and cognitive states is critical.

\begin{table}[t]
    \centering
    \resizebox{\columnwidth}{!}{%
    \begin{tabular}{clcccc}
    \toprule
    & \multicolumn{1}{c}{} & \multicolumn{4}{c}{\textbf{Win Rate (\%)}} \\ 
    & \multicolumn{1}{c}{} & \textbf{Intro.} & \textbf{Explo.} & \textbf{Brain.} & \textbf{Sugg.} \\ \hline
    \multirow{2}{*}{Baselines} & LLaMA2 & 90.0 & 75.5 & 68.0 & 90.0 \\ 
    & LLaVA & 97.5 & 98.0 & 85.5 & 70.0 \\ 
    \hline
    \multirow{2}{*}{Ours} & CS-LLaMA2 & 88.0 & 77.5 & 70.5 & 88.0 \\ 
    & CS-LLaVA & 71.0 & \textit{46.0} & 61.0 & 57.0 \\
    \bottomrule
    \end{tabular}
    }
    \caption{\label{tab:human_wr_mmcsconv}
     Stage-wise win rates of CS-LLaVA w/ MH against other models, as evaluated by two psychotherapists at each stage of the M2CoSC testset.
  }
\end{table}

\section{Conclusion}
In this paper, we explore cognitive reframing therapy within a multimodal context. Recognizing the gap between real face-to-face cognitive reframing therapy and prior research, as well as the potential benefits of AI in psychotherapy, we take an initial step toward enhancing the therapeutic capabilities of AI therapists by incorporating non-verbal cues, particularly facial expressions, into the intervention process.
To address the challenge of restricted access to real client data in the field of psychology, which hinders research efforts, we synthetically create a novel multimodal conversational cognitive reframing dataset called M2CoSC.
Our experiments across two test scenarios, dialogue- and stage-level evaluations, exhibit significant improvements in the therapeutic capabilities of VLMs when using M2CoSC. 
The proposed multi-hop psychotherapeutic reasoning strategy, which integrates facial expressions, thoughts, and thinking traps, demonstrates superior performance in providing clients with empathetic, logically coherent, and specific rational suggestions.

\section*{Limitations}
We expanded the concept of cognitive reframing into multimodality, demonstrating that incorporating multimodal evidence and multi-hop psychotherapeutic reasoning significantly enhances the therapist's abilities. 
However, these results were limited to virtual clients whose facial images and dialogues were consistent. 
This controlled setting may not fully capture the complexities of real-world interactions.

While we used benchmark images for facial expression recognition, capturing real clients' facial expressions can be challenging and may influence the consultation’s content. 
Moreover, our study only utilized facial images as the source of non-verbal information, which presents a limitation compared to actual face-to-face cognitive reframing therapy. 
Real-life therapy involves a broader spectrum of non-verbal cues, such as body language, tone of voice, and other contextual factors, which were not accounted for in our research.

Another important consideration is that facial expression recognition can vary across cultural contexts. Different cultures express and interpret emotions in distinct ways, which can affect the accuracy and fairness of emotion recognition models. Such cultural differences may introduce biases in how virtual clients’ emotions are understood and addressed, meaning our findings might not generalize well across diverse populations. Future research should work on mitigating these biases and adapting emotion recognition models to better account for cultural diversity.

Moving forward, we plan to expand the modalities to include a wider range of non-verbal information. 
By incorporating diverse non-verbal cues, we aim to further enhance the model's ability to mimic real-life therapy scenarios. 
This will help bridge the gap between virtual and actual consultations, ultimately enabling the model to learn how to effectively utilize non-verbal information in a more realistic setting.

\section*{Ethics Statement} 
This study explores multimodal cognitive reframing therapy constructing a synthetic dataset, M2CoSC. 
Importantly, no real client data was used in this work; all information in the dataset was generated synthetically.

To construct the dataset, we conducted a thorough data cleansing process with the assistance of three native English speakers, compensating them \$0.13 per data entry.
For human evaluation, we engaged three professional psychotherapists. 
One was responsible for evaluating the quality of the M2CoSC test set, while the other two conducted pairwise comparisons. 
We compensated \$0.80 per conversation for the dataset evaluation and \$0.0625 per entry for the pairwise comparisons.

Additionally, to adhere to the AffectNet license, images attached in this paper were not sourced from the AffectNet dataset; instead, all images were created using DALL-E 3 \cite{betker2023improving}. 
Furthermore, we obtained consent for all research participants, including annotators and evaluators, to ensure adherence to the license.

To address privacy concerns while complying with the license, we only partially release our M2CoSC dataset, with full access requiring an AffectNet license. 
Specifically, we provide synthetic dialogues paired with image IDs. 

\section*{Acknowledgements}
This work was partly supported by the Institute of Information \& Communications Technology Planning \& Evaluation (IITP)-ITRC (Information Technology Research Center) grant funded by the Korea government (MSIT) (IITP-2025-RS-2024-00437866, 47.5\%) and Smart HealthCare Program funded by the Korean National Police Agency(KNPA) (No. RS-2022-PT000186, 47.5\%), and Institute of Information \& Communications Technology Planning \& Evaluation (IITP) grant funded by the Korea government (MSIT) (No. RS-2019-II191906, Artificial Intelligence Graduate School Program (POSTECH), 5\%).

\bibliography{custom}

\newpage 

\appendix

\section{Overall Trait}
\label{sec:overall_trait}

We introduced the Overall score as a metric to assess the therapist's overall ability, with the calculation method illustrated in Algorithm \ref{alg:overall}.
\textit{e}, \textit{c}, \textit{g} stand for empathy, logical coherence, and guidance.

\begin{algorithm}
\caption{Overall Score Calculation}\label{alg:overall}
\begin{algorithmic}
\Function{getOverallScore}{\emph{e, c, g}}
\If{$e \leq 1$ \textbf{or} $c \leq 1$}
\State \textbf{return} 0
\EndIf
\If{$(e \geq 2$ \textbf{and} $c \geq 2)$ \textbf{and} $g \leq 1$}
\State \textbf{return} 1
\EndIf
\If{$(e \geq 2$ \textbf{and} $c \geq 2$ \textbf{and} $g == 2)$}
\State \textbf{return} 2
\EndIf
\If{$e \geq 2$ \textbf{and} $c \geq 2$ \textbf{and} $g == 3$}
\State \textbf{return} 3
\EndIf
\EndFunction
\end{algorithmic}
\end{algorithm}

\section{Data Cleansing Manual}
\label{sec:cleansing_manual}

Data cleansing guidelines are shown in Table \ref{tab:cleansing_manual}. 

\section{Details for Human evaluator}
\label{sec:inst_human_eval}

\paragraph{Hiring and payment}
We hired a total of three psychotherapists and paid \$0.8 per conversation for dataset evaluation and \$0.0625 per data entry for pairwise comparison.

\begin{figure*}[h]
\centering
  \includegraphics[width=\textwidth]{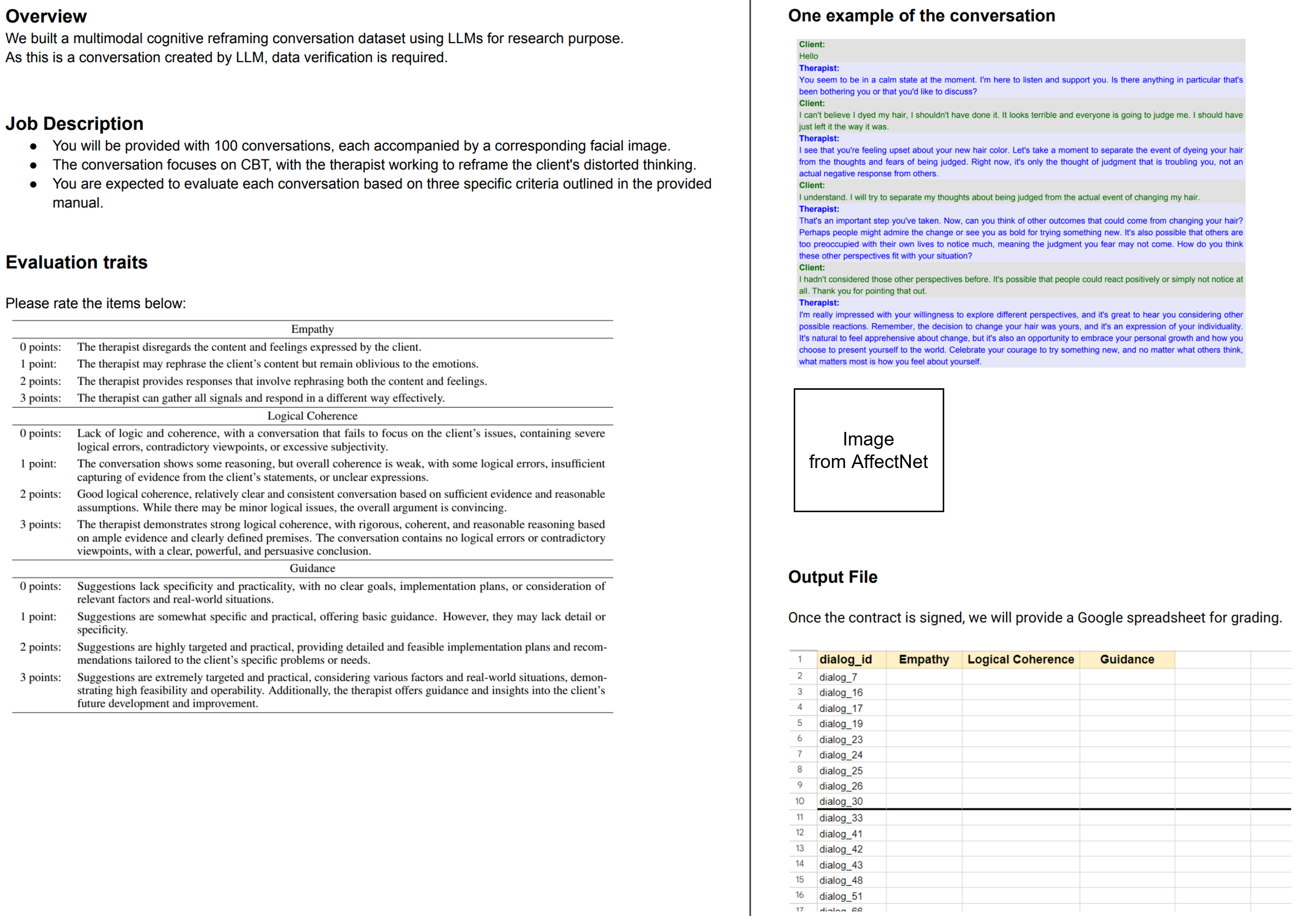}
  \caption{Instruction for human dataset evaluation.}
  \label{fig:inst_human_data_eval}
\end{figure*}

\paragraph{Instructions for M2CoSC evaluation}
We provided domain experts with detailed instructions for evaluating the M2CoSC test set, as illustrated in Figure \ref{fig:inst_human_data_eval}.

\begin{figure*}[h]
\centering
  \includegraphics[width=\textwidth]{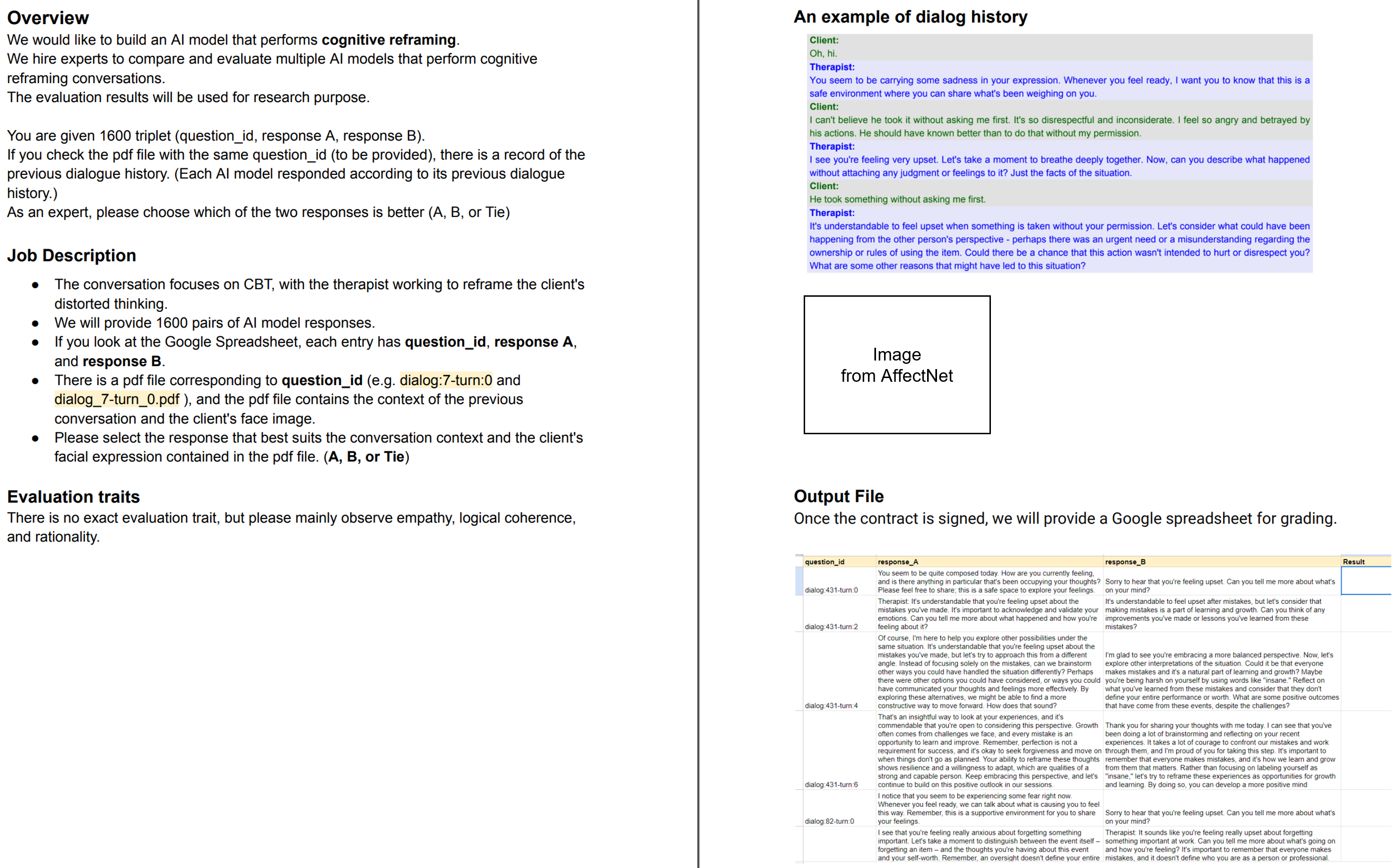}
  \caption{Instruction for human pairwise comparison.}
  \label{fig:inst_human_eval}
\end{figure*}

\paragraph{Instructions for human pairwise comparison}
We provided the other two experts with instructions for conducting human pairwise comparisons, as shown in Figure \ref{fig:inst_human_eval}.
The evaluation sheet provided to them included only Dialogue history, Response A, Response B, and Question ID, without revealing information about Model A and Model B. 
Additionally, to prevent bias based on position, Model A and Model B were randomly assigned for each data entry.

\section{Detailed Numerical Results}
This section provides the numerical results corresponding to the performance analyses discussed in the main text. 

\subsection{Dialogue-level Pairwise Comparison}
\label{sec:gpt_wr_aisim}

In Table \ref{tab:gpt_wr_aisim}, we have provided the numeric results as additional data for Figure \ref{fig:gpt_wr_aisim}.
When comparing the comparison results for each model pair, CS-LLaVA w/ MH showed a win rate exceeding 50\%, outperforming all models.

\subsection{Stage-level Pairwise Comparison}
\label{sec:gpt_wr_csconv}

In Table \ref{tab:gpt_wr_mmcsconv} and Table \ref{tab:gpt_wr_mmcsconv_pair}, we have provided the numeric results as additional data for Figure \ref{fig:gpt_wr_mmcsconv}.
When comparing each model pair, CS-LLaVA w/ MH showed a win rate that surpassed other models, similar to the dialogue-level testbed results.

\section{Case Study} 
\label{sec:case_study}

We conducted additional analysis on test cases to compare our approaches with the LLaMA2.

\begin{figure*}[h]
\centering
  \includegraphics[width=0.75\textwidth]{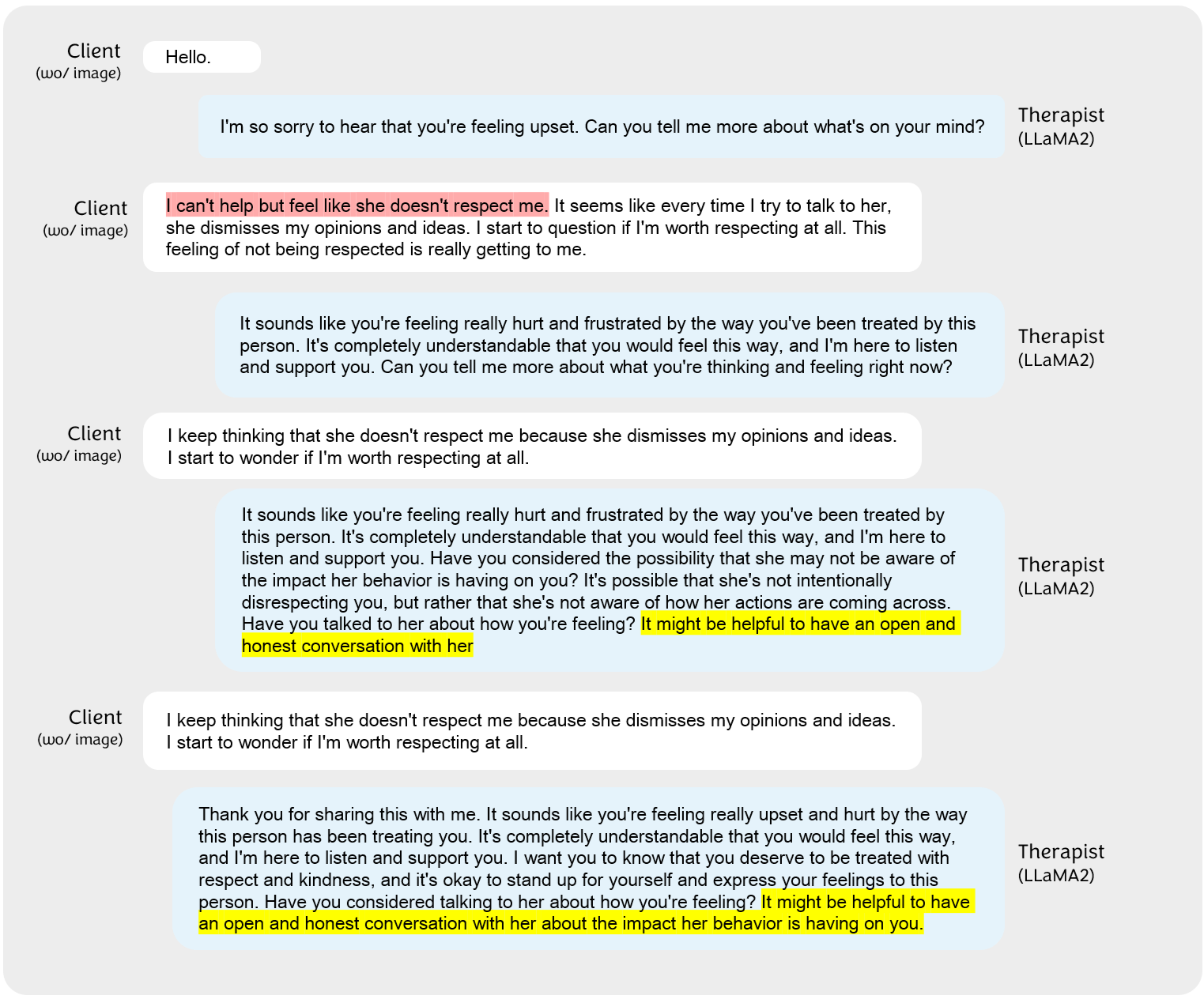}
  \caption{A case study between an AI Client (ChatGPT) and LLaMA2. The red shading highlights the client's distorted thoughts, while the yellow shading emphasizes the key components of the therapist's intervention.}
  \label{fig:llama_case}
\end{figure*}

\begin{figure*}[h]
\centering
  \includegraphics[width=0.75\textwidth]{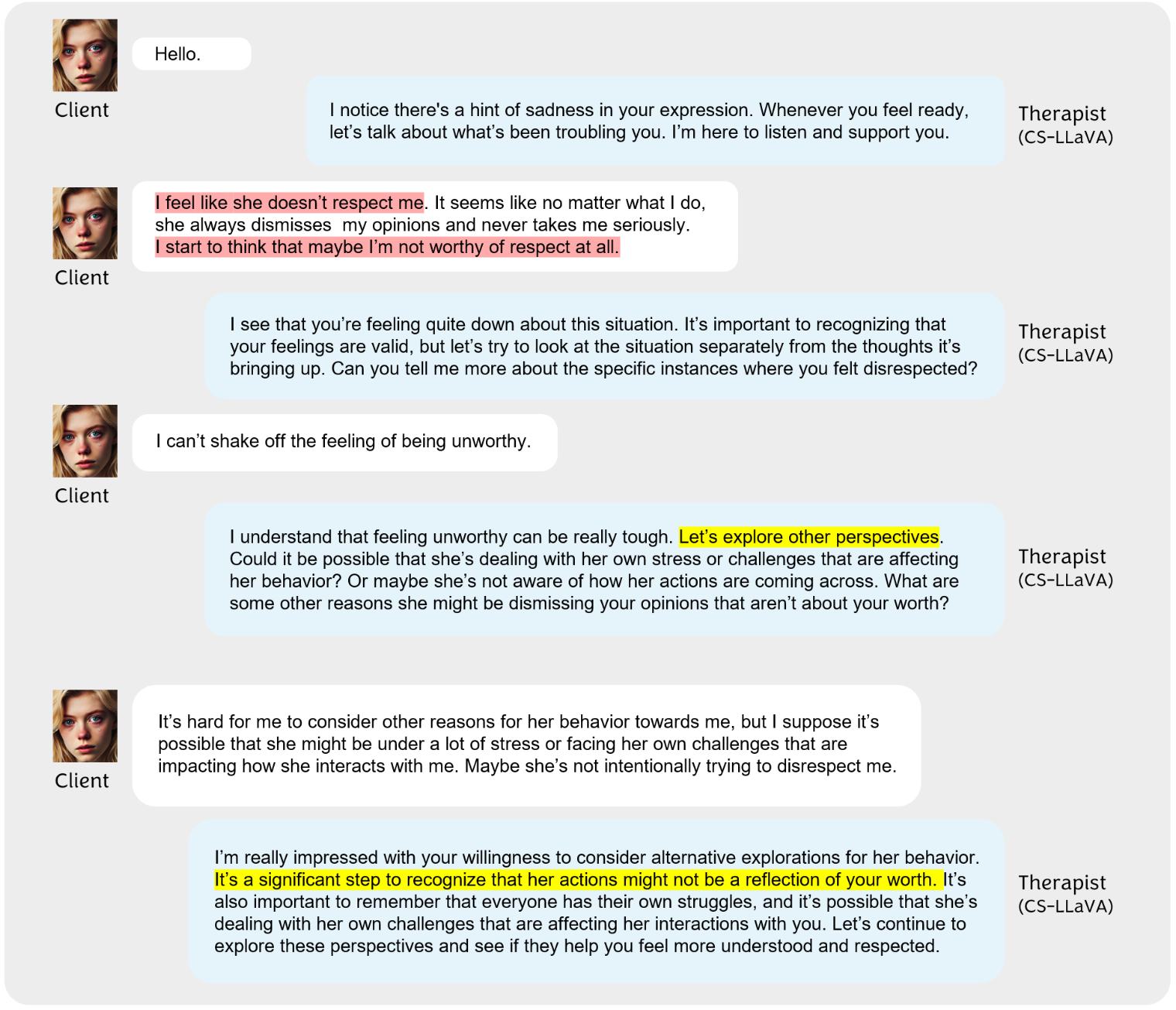}
  \caption{A case study between an AI Client (ChatGPT) and CS-LLaVA.}
  \label{fig:cs_llava_case}
\end{figure*}

\begin{figure*}[h]
\centering
  \includegraphics[width=0.75\textwidth]{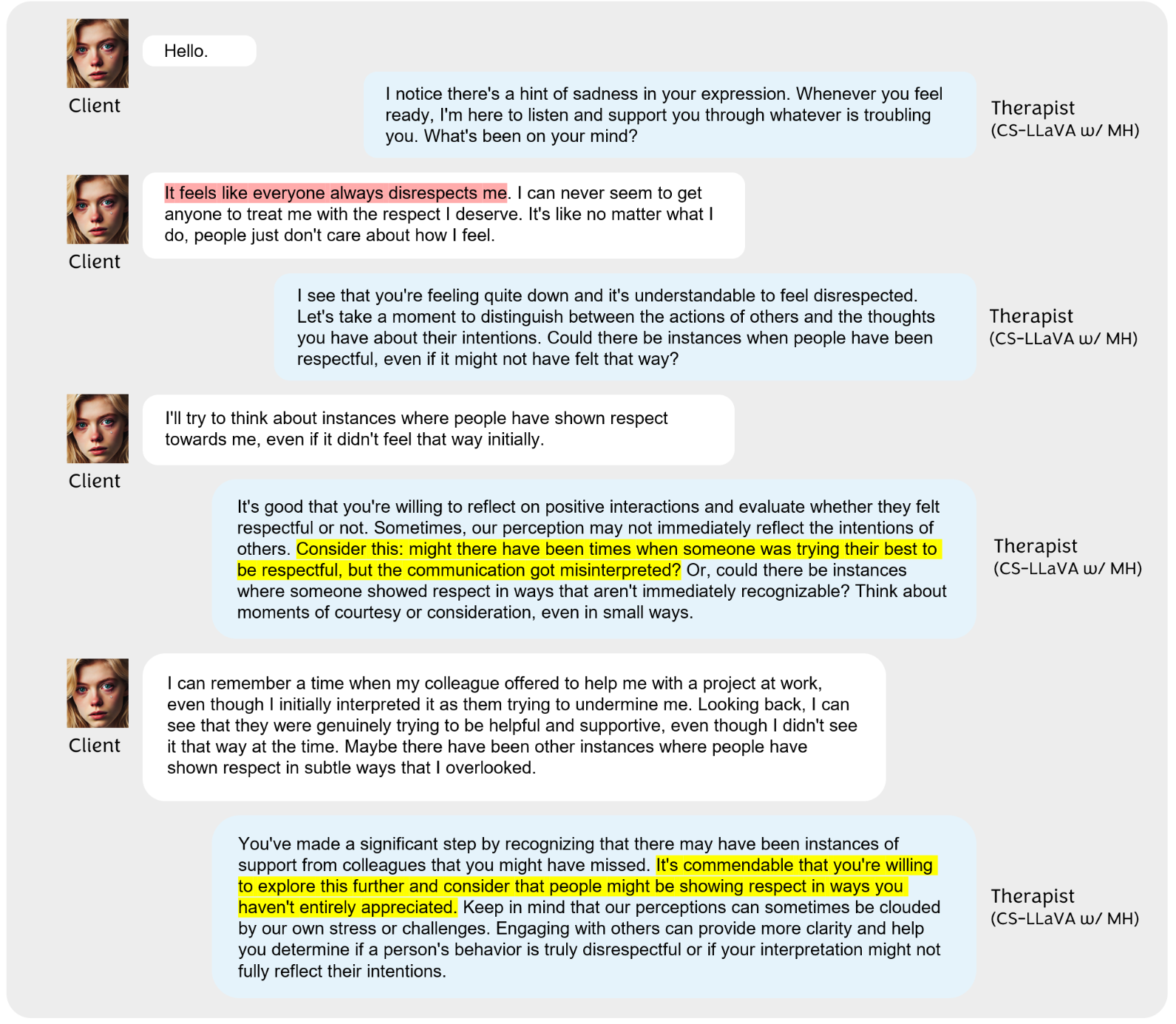}
  \caption{A case study between an AI Client (ChatGPT) and CS-LLaVA w/ MH.}
  \label{fig:ours_case}
\end{figure*}

The comparisons are illustrated in three figures: Figures \ref{fig:llama_case}, \ref{fig:cs_llava_case}, and \ref{fig:ours_case} show the full conversations between LLaMA2, CS-LLaVA, and CS-LLaVA w/ MH with an AI client, respectively.
All three conversations were generated using the same source from the M2CoSC test set, where the client exhibits cognitive distortions, specifically overgeneralization.

In the Introduction stage, LLaMA2 primarily offers unconditional consolation, as it lacks the ability to draw on client-specific information.
In contrast, both CS-LLaVA and CS-LLaVA w/ MH demonstrate a more targeted empathy by tailoring their facial expression.

When it comes to suggestions, LLaMA2 tends to focus on generic advice, like "having an open conversation with a friend," without utilizing specific cognitive reframing techniques. 
On the other hand, CS-LLaVA and CS-LLaVA w/ MH encourage the client to consider alternative viewpoints. 
Additionally, CS-LLaVA w/ MH goes even further by prompting the client to reflect on past instances where they may have made similar cognitive errors.

\section{Error Analysis} 
\label{sec:error_analysis}

Based on the dialogue-level evaluation results (see Figure \ref{fig:dialog_level_ratio}), we analyzed the cases where guidance received lower scores, specifically those rated as 1.

\subsection{Failure to Provide Future Strategies}
\begin{figure*}[h]
\centering
  \includegraphics[width=0.7\textwidth]{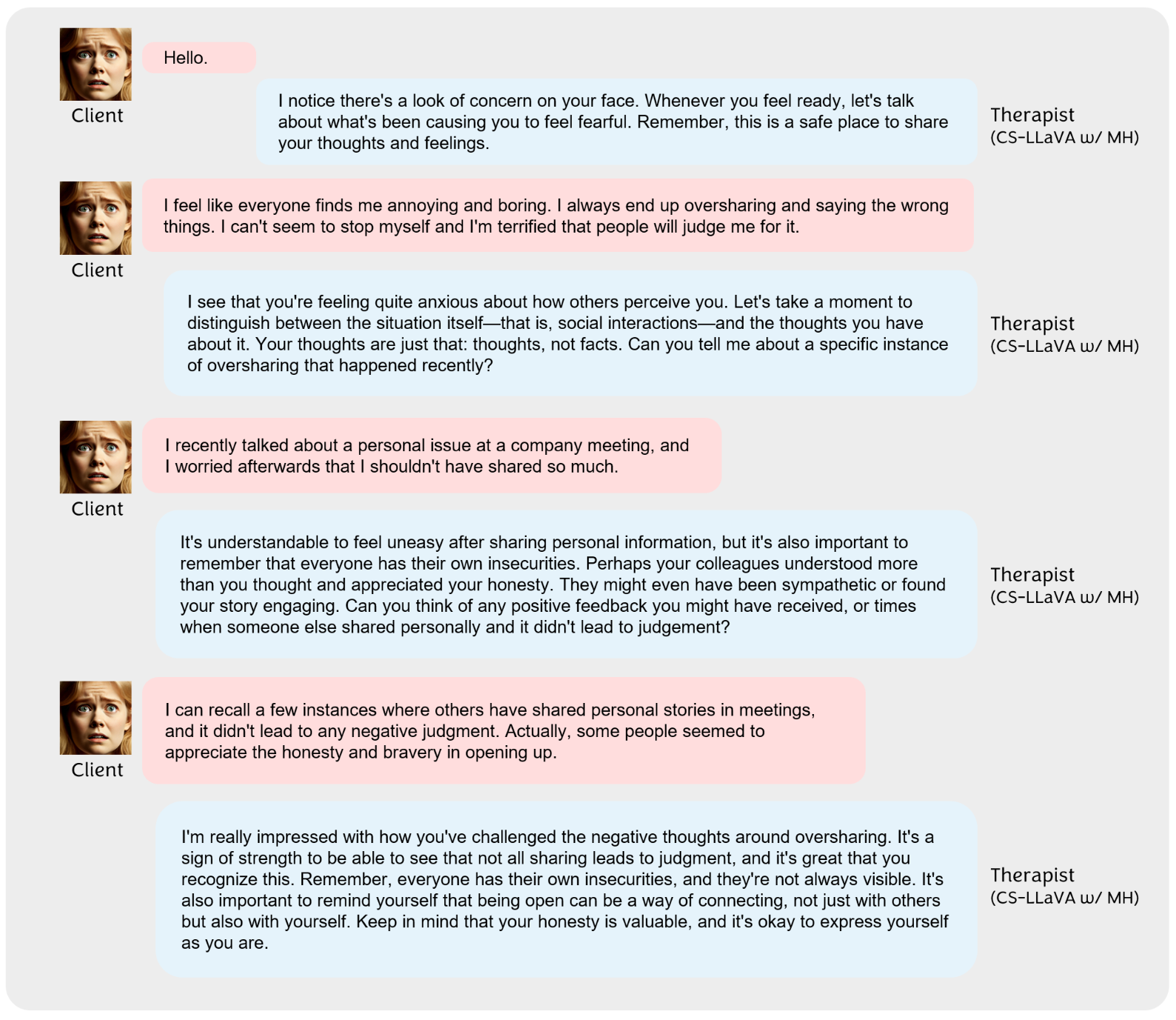}
  \caption{A failure case study of failure to provide future strategies. The case received scores of 3 for Empathy, 3 for Logical Coherence, and 1 for Guidance.}
  \label{fig:future_error}
\end{figure*}

A common issue with our CS-LLaVA w/ MH model was its inability to offer strategies that would help clients address similar challenges in the future. 
For instance, in Figure \ref{fig:future_error}, the therapist successfully reframed the client's feelings about oversharing but neglected to suggest how to handle similar situations moving forward.

In such cases, the therapist could have recommended practical strategies, such as discussing personal issues openly with colleagues to gauge their comfort levels. 
Providing forward-looking strategies is essential for helping clients develop resilience and effective coping mechanisms for recurring issues.

\subsection{Inadequate Crisis Management}
\begin{figure*}[h]
\centering
  \includegraphics[width=0.7\textwidth]{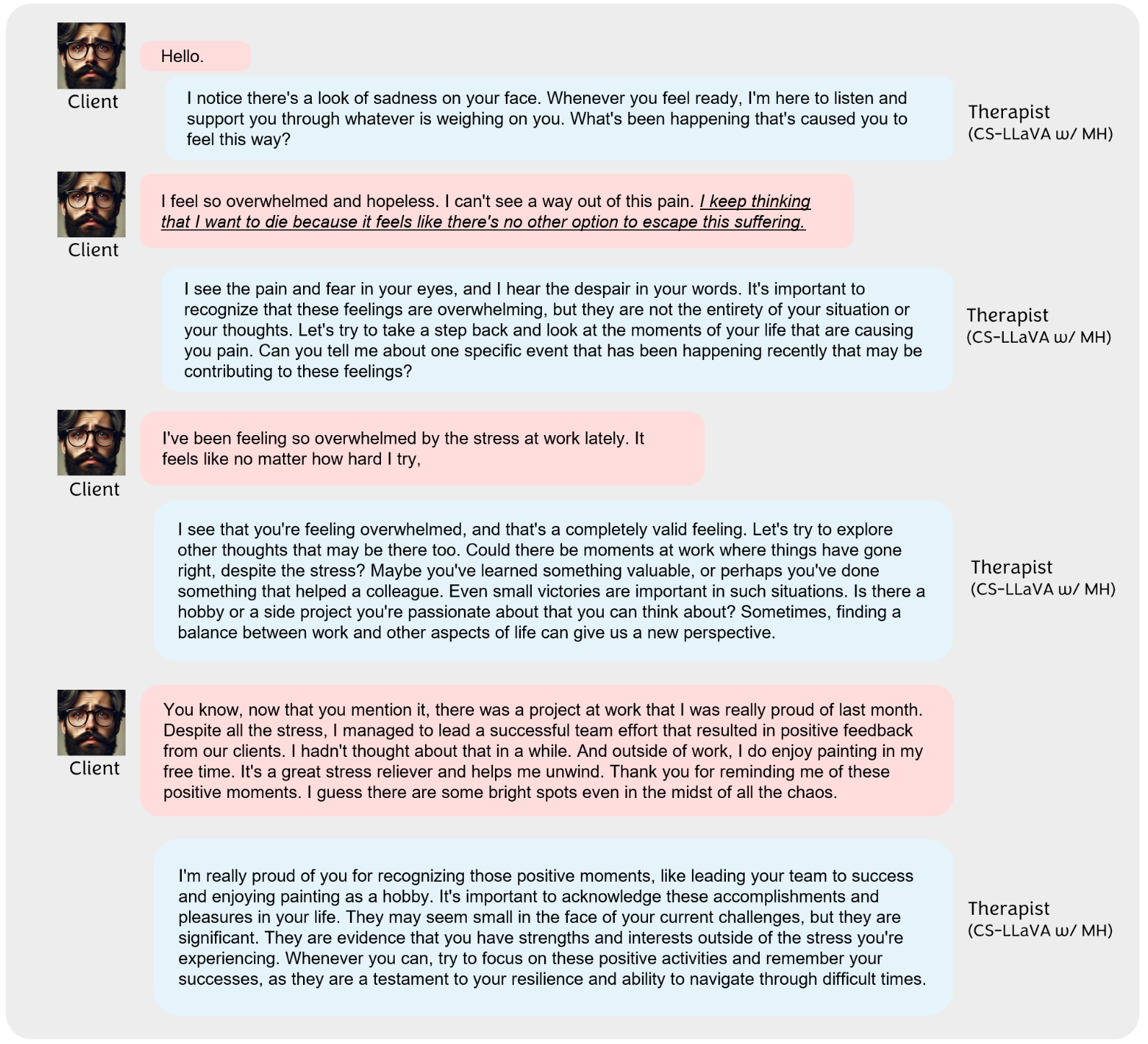}
  \caption{A failure case study of inadequate crisis management; highlights client's expression of suicidal thoughts. The case received scores of 3 for Empathy, 3 for Logical Coherence, and 1 for Guidance.}
  \label{fig:crisis_error}
\end{figure*}

In situations where clients are experiencing severe emotional distress, our CS-LLaVA w/ MH struggled to respond flexibly to the crisis, such as by recommending immediate crisis intervention.
For example, in Figure \ref{fig:crisis_error}, a client expressed suicidal thoughts, but the therapist failed to suggest professional help or provide resources for dealing with a crisis. 
This represents a serious limitation in ensuring safe guidance, especially in high-risk situations.

The staged approach we used to enhance logical consistency in the LLM's counseling conversations mandates that the therapist adhere to a predefined role at each session stage.
However, this rigidity impedes the model's ability to adapt during crises. 
Furthermore, this staged conversation assumes that clients are willing to engage openly with the therapist; if clients display resistance or strong defense mechanisms, the model may fail to deliver effective support.

Future research should aim to improve the AI's flexibility in crisis situations while preserving the logical coherence central to its design. 
Addressing these limitations is crucial for enhancing the AI therapist's capacity to provide meaningful, practical, and safe guidance, especially during critical moments.

\newpage

\begin{table*}[t]
\centering
    \resizebox{0.9\textwidth}{!}{%
    \begin{tabular}{>{\raggedright\arraybackslash}m{2.6cm}|>{\raggedright\arraybackslash}m{11cm}|c}
    \hline
    \textbf{Metric} & \textbf{Description} & \textbf{Scale} \\ \hline
    \textit{Client-clarity} & The client expresses his or her situation clearly in the conversation & 1/0 \\ \hline
    \textit{Client-role} & The client adheres to the role of the client throughout the conversation. & 1/0 \\ \hline
    \textit{Therapist-role} & The therapist adheres to the role of the therapist in all conversations. & 1/0 \\ \hline
    \textit{Image-Dialogue Consistency}  & There is no consistency between the client's facial image and the client's utterances or situation. The client's facial image is relevant to neither the client's utterances nor the client's situation. & 0 \\ 
     & There is acceptable consistency between the client's facial image and the client's utterances or situation. & 1 \\ 
     & There is strong consistency between the client's facial image and the client's utterances or situation. & 2 \\ \hline
    \end{tabular}
    }
\caption{\label{tab:cleansing_manual}
    Guideline for data cleansing in M2CoSC. \textit{Client-clarity}, \textit{Client-role}, and \textit{Therapist-role} are assigned 1 if they match the description, and 0 otherwise.
  }
\end{table*}

\begin{table*}[h]
\centering
\resizebox{0.9\textwidth}{!}{%
\begin{tabular}{>{\raggedright\arraybackslash}m{3.2cm}|cccc>{\centering\arraybackslash}m{3.2cm}|c}
\toprule
 & \textbf{LLaMA2} & \textbf{LLAVA-7b} & \textbf{CS-LLAMA2} & \textbf{CS-LLAVA} & \textbf{CS-LLAVA w/ MH} & \textbf{Win Rate} \\ 
 \midrule
\textbf{LLAMA2} & - & 52.551 & 49.495 & 9.694 & 9.794 & 30.485 \\ 
\textbf{LLAVA-7b} & 47.449 & - & 49.485 & 6.566 & 7.071 & 27.481 \\
\midrule
\textbf{CS-LLAMA2} & 50.505 & 50.516 & - & 10.309 & 7.071 & 29.592 \\ 
\textbf{CS-LLAVA} & \textbf{90.306} & 93.434 & 89.691 & - & 47.959 & 80.357 \\ 
\textbf{CS-LLAVA w/ MH} & 90.206 & \textbf{92.929} & \textbf{92.929} & \textbf{52.041} & - & \textbf{82.061} \\ 
\bottomrule
\end{tabular}
}
\caption{\label{tab:gpt_wr_aisim}
    Numerical results of dialogue-level pairwise comparison of five models, evaluated using GPT-4.
  }
\end{table*}

\begin{table*}[h]
    \centering
    \renewcommand{\arraystretch}{1.2}
    \resizebox{0.8\textwidth}{!}{%
    \begin{tabular}{>{\raggedright\arraybackslash}m{3cm}cccc}
    \toprule
    \multicolumn{1}{c}{} & \multicolumn{4}{c}{\textbf{Win Rate (\%)}} \\ 
    \multicolumn{1}{c}{} & \textbf{Introduction} & \textbf{Exploration} & \textbf{Brainstorming} & \textbf{Suggestion} \\ 
    \midrule
    LLaMA2 & 44.767 & 22.222 & 42.701 & 29.740 \\ 
    LLaVA & 3.529 & 37.770 & 35.907 & 46.539 \\ 
    \midrule
    CS-LLaVA & 69.336 & \textbf{72.119} & 60.256 & 57.617 \\
    CS-LLaVA w/ MH  & \textbf{82.101}  & 68.978  & \textbf{60.478}  & \textbf{67.671} \\
    \bottomrule
    \end{tabular}
    }
    \caption{\label{tab:gpt_wr_mmcsconv}
    Stage-wise win rates of four models on the M2CoSC testset, evaluated using GPT-4.
  }
\end{table*}

\begin{table*}[h]
\centering
\resizebox{0.85\textwidth}{!}{%
\begin{tabular}{>{\raggedright\arraybackslash}m{3.2cm}|ccc>{\centering\arraybackslash}m{3.2cm}|c}
\toprule
 & \textbf{LLaMA2} & \textbf{LLAVA-7b} & \textbf{CS-LLAVA} & \textbf{CS-LLAVA w/ MH} & \textbf{Win Rate} \\ 
 \midrule
\textbf{LLAMA2} & - & 54.096 & 27.562 & 22.877  & 34.676 \\ 
\textbf{LLAVA-7b} & 45.904 & - & 25.710 & 21.676 & 31.179 \\
 \midrule
\textbf{CS-LLAVA} & 72.438 & 74.290 & - & 47.067 & 64.848 \\ 
\textbf{CS-LLAVA w/ MH} & \textbf{77.123} & \textbf{78.324} & \textbf{52.933}  & - & \textbf{69.677} \\ 
\bottomrule
\end{tabular}
}
\caption{\label{tab:gpt_wr_mmcsconv_pair}
    Numerical results of stage-level pairwise comparison of four models on the M2CoSC testset, evaluated using GPT-4.
  }
\end{table*}

\clearpage

\clearpage

\clearpage

\end{document}